%% file: emnlp-ijcnlp-2019.tex
\documentclass[11pt,a4paper]{article}
\usepackage[hyperref]{emnlp-ijcnlp-2019}
\usepackage{times}
\usepackage{latexsym}
\usepackage{amssymb}
\usepackage{multirow}
\usepackage{booktabs}
\usepackage{amsmath}
\usepackage{graphicx}
\usepackage{pifont}
\newcommand{\cmark}{\ding{51}}%
\newcommand{\xmark}{\ding{55}}%
\usepackage{url}
\usepackage{lipsum}

\aclfinalcopy 

\title{Aspect-based Sentiment Classification with Aspect-specific Graph Convolutional Networks}

\author{
  Chen Zhang \\
  Beijing Institute of Technology \\
  Beijing, China \\
  {\tt gene@bit.edu.cn} \\\And
  Qiuchi Li \\
  University of Padua \\
  Padua, Italy \\
  {\tt qiuchili@dei.unipd.it} \\\And
  Dawei Song\Thanks{ Corresponding author.} \\
  Beijing Institute of Technology \\
  Beijing, China \\
  {\tt dwsong@bit.edu.cn} \\
 }
 
\date{}

\begin{document}
\maketitle
\begin{abstract}
Due to their inherent capability in semantic alignment of aspects and their context words, attention mechanism and Convolutional Neural Networks (CNNs) are widely applied for aspect-based sentiment classification. However, these models lack a mechanism to account for relevant syntactical constraints and long-range word dependencies, and hence may mistakenly recognize syntactically irrelevant contextual words as clues for judging aspect sentiment. To tackle this problem, we propose to build a Graph Convolutional Network (GCN) over the dependency tree of a sentence to exploit syntactical information and word dependencies. Based on it, a novel aspect-specific sentiment classification framework is raised. Experiments on three benchmarking collections illustrate that our proposed model has comparable effectiveness to a range of state-of-the-art models\footnote{Code and preprocessed datasets are available at \href{https://github.com/GeneZC/ASGCN}{https://github.com/GeneZC/ASGCN}.}, and further demonstrate that both syntactical information and long-range word dependencies are properly captured by the graph convolution structure.
\end{abstract}

\input{introduction}
\input{background}
\input{model}
\input{experiment}

\input{discussion}

\input{related}
\input{conclusion}

\section*{Acknowledgments}

This work is supported by The National Key Research and Development Program of China (grant No. 2018YFC0831704), Natural Science Foundation of China (grant No. U1636203, 61772363), Major Project of Zhejiang Lab (grant No. 2019DH0ZX01), and the European Union’s Horizon 2020 Research and Innovation Programme under the Marie Skłodowska-Curie grant agreement No. 721321.

\bibliography{emnlp-ijcnlp-2019}
\bibliographystyle{acl_natbib}



\end{document}

%% file: introduction.tex
\section{Introduction}

Aspect-based (also known as aspect-level)  sentiment classification aims at identifying the sentiment polarities of aspects explicitly given in sentences. For example, in a comment about a laptop saying ``\textit{From the speed to the multi-touch gestures this operating system beats Windows easily.}'', the sentiment polarities for two aspects \textit{operating system} and \textit{Windows} are \textit{positive} and \textit{negative}, respectively. Generally, this task is formulated as predicting the polarity of a provided (sentence, aspect) pair.

Given the inefficiency of manual feature refinement~\cite{jiang2011target}, early works of aspect-based sentiment classification are mainly based on neural network methods~\cite{dong2014adaptive,vo2015target}. Ever since~\citet{tang2016effective} pointed out the challenge of modelling semantic relatedness between context words and aspects, attention mechanism coupled with Recurrent Neural Networks (RNNs)~\cite{bahdanau2014neural,luong2015effective,xu2015show} starts to play a critical role in more recent models~\cite{wang2016attention,tang2016aspect,yang2017attention,liu2017attention,ma2017interactive,huang2018aspect}.

While attention-based models are promising, they are insufficient to capture syntactical dependencies between context words and the aspect within a sentence. Consequently, \textbf{the current attention mechanism may lead to a given aspect mistakenly attending to syntactically unrelated context words as descriptors} (Limitation 1). Look at a concrete example ``\textit{Its size is ideal and the weight is acceptable.}''. Attention-based models often identify \textit{acceptable} as a descriptor of the aspect \textit{size}, which is in fact not the case. In order to address the issue, ~\citet{he2018effective} imposed some syntactical constraints on attention weights, but the effect of syntactical structure was not fully exploited.

In addition to the attention-based models, Convolutional Neural Networks (CNNs)~\cite{xue2018aspect,li2018transformation} have been employed to discover descriptive multi-word phrases for an aspect, based on the finding~\cite{fan2018convolution} that the sentiment of an aspect is usually determined by key phrases instead of individual words. Nevertheless, the CNN-based models can only perceive multi-word features as consecutive words with the convolution operations over word sequences, but are  \textbf{inadequate to determine sentiments depicted by multiple words that are not next to each other} (Limitation 2). In the sentence ``\textit{The staff should be a bit more friendly}'' with \textit{staff} as the aspect, a CNN-based model may make an incorrect prediction by detecting \textit{more friendly} as the descriptive phrase, disregarding the impact of \textit{should be} which is two words away but reverses the sentiment. 

In this paper, we aim to tackle the two limitations identified above by using Graph Convolutional Networks (GCNs)~\cite{kipf2017semi}. GCN has a multi-layer architecture, with each layer encoding and updating the representation of nodes in the graph using features of immediate neighbors. Through referring to syntactical dependency trees, a GCN is potentially capable of drawing syntactically relevant words to the target aspect, and exploiting long-range multi-word relations and syntactical information with GCN layers. GCNs have been deployed on document-word relationships~\cite{yao2018graph} and tree structures~\cite{marcheggiani2017encoding,zhang2018graph}, but how they can be effectively used in aspect-based sentiment classification is yet to be explored. 

To fill the gap, this paper proposes an Aspect-specific Graph Convolutional Network (ASGCN), which, to the best of our knowledge, is the first GCN-based model for aspect-based sentiment classification. ASGCN starts with a bidirectional Long Short-Term Memory network (LSTM) layer to capture contextual information regarding word orders. In order to obtain aspect-specific features, a multi-layered graph convolution structure is implemented on top of the LSTM output, followed by a masking mechanism that filters out non-aspect words and keeps solely high-level aspect-specific features. The aspect-specific features are fed back to the LSTM output for retrieving informative features with respect to the aspect, which are then used to predict aspect-based sentiment. 

Experiments on three benchmarking datasets show that ASGCN effectively addresses both limitations of the current aspect-based sentiment classification approaches, and outperforms a range of state-of-the-art models. 

Our contributions are as follows:
\begin{itemize}
    \item We propose to exploit syntactical dependency structures within a sentence and resolve the long-range multi-word dependency issue for aspect-based sentiment classification.
    
    \item We posit that Graph Convolutional Network (GCN) is suitable for our purpose, and propose a novel Aspect-specific GCN model. To our best knowledge, this is the first investigation in this direction.
    
    \item Extensive experiment results verify the importance of leveraging syntactical information and long-range word dependencies, and demonstrate the effectiveness of our model in capturing and exploiting them in aspect-based sentiment classification.
\end{itemize}

%% file: background.tex
\section{Graph Convolutional Networks}

\begin{figure}[]
    \centering
    \includegraphics[scale=0.5]{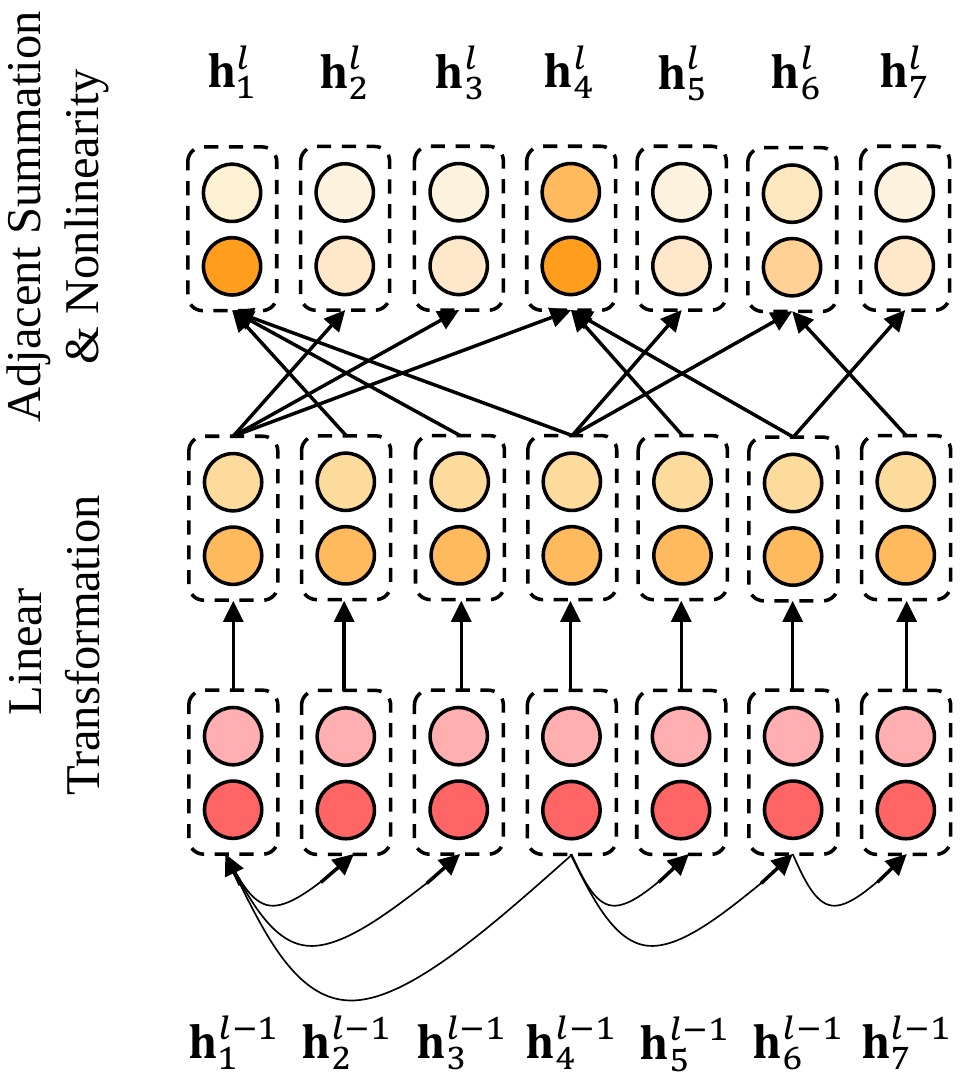}
    \caption{\label{fig1} An example of GCN layer.}
\end{figure}

GCNs can be considered as an adaptation of the conventional CNNs for encoding local information of unstructured data. For a given graph with $k$ nodes, an adjacency matrix\footnote{$\mathbf{A}_{ij}$ indicates whether the $i$-th token is adjacent to the $j$-th token or not.} $\mathbf{A}\in\mathbb{R}^{k\times k}$ is obtained through enumerating the graph. For convenience, we denote the output of the $l$-th layer for node $i$ as $\mathbf{h}_i^{l}$, where $\mathbf{h}_i^{0}$ represents the initial state of node $i$.  For an $L$-layer GCN, $l\in [1,2,\cdots,L]$ and $\mathbf{h}_i^{L}$ is the final state of node $i$. The graph convolution operated on the node representation can be written as:
\begin{equation}
    \mathbf{h}_i^{l}=\sigma(\sum_{j=1}^{k}\mathbf{A}_{ij}\mathbf{W}^l\mathbf{h}_j^{l-1}+\mathbf{b}^l)
\end{equation}
where $\mathbf{W}^{l}$ is a linear transformation weight,  $\mathbf{b}^l$ is a bias term, and $\sigma$ is a nonlinear function, e.g. ReLU. For a better illustration, an example of GCN layer is shown in Figure~\ref{fig1}.

As the graph convolution process only encodes information of immediate neighbors, a node in the graph can  only be influenced by the neighbouring nodes within $L$ steps in an $L$-layer GCN. In this way, the graph convolution over the dependency tree of a sentence provides syntactical constraints for an aspect within the sentence to identify descriptive words based on syntactical distances. Moreover, GCN is able to deal with the circumstances where the polarity of an aspect is described by non-consecutive words, as GCN over dependency tree will gather the non-consecutive words into a smaller scope and aggregate their features properly with graph convolution. Therefore, we are inspired to adopt GCN to leverage syntactical information and long-range word dependencies for aspect-based sentiment classification.

%% file: model.tex
\section{Aspect-specific Graph Convolutional Network}

\begin{figure*}[]
    \centering
    \includegraphics[scale=0.5]{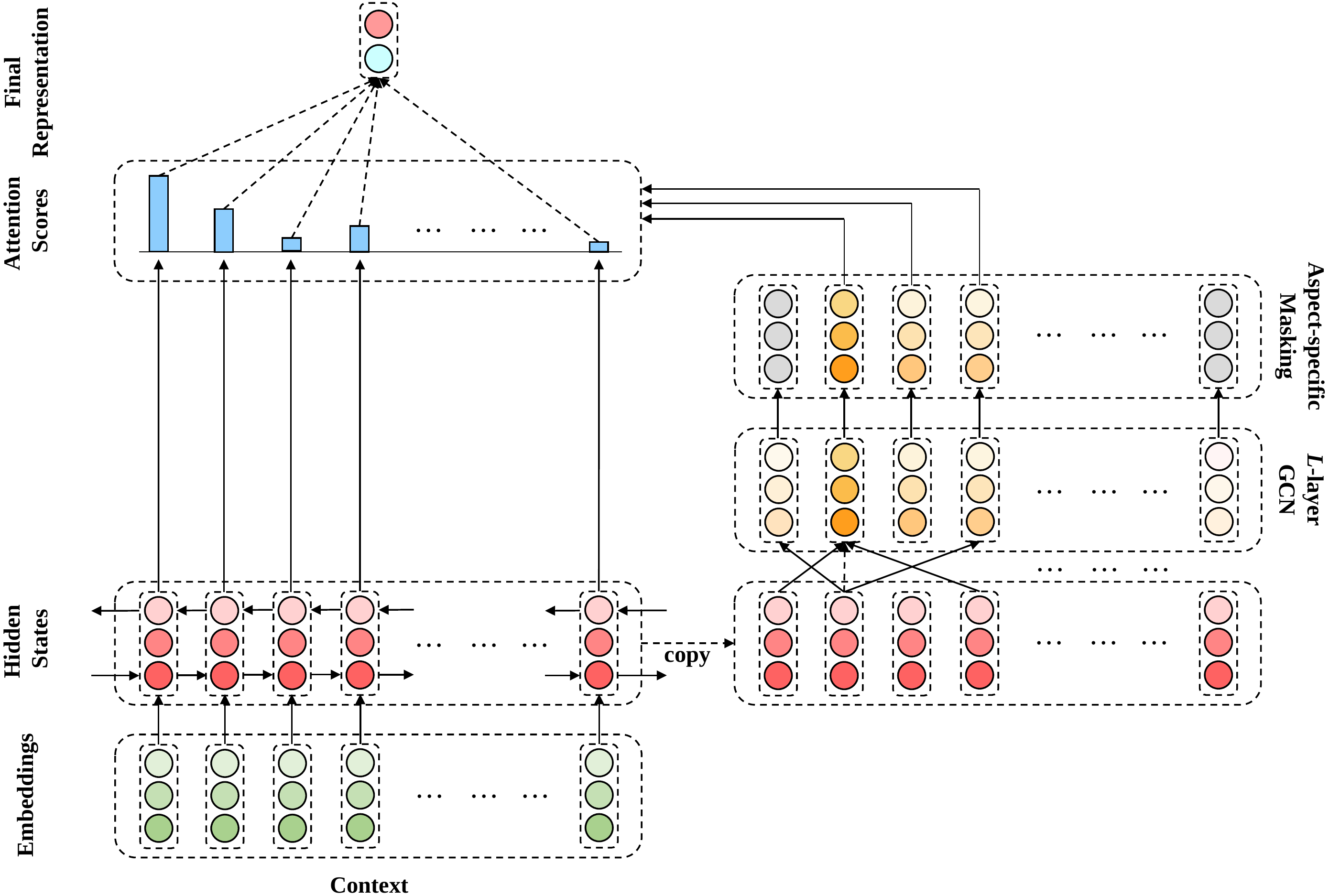}
    \caption{\label{fig2} Overview of aspect-specific graph convolutional network.}
\end{figure*}

Figure~\ref{fig2} gives an overview of ASGCN. The components of ASGCN will be introduced separately in the rest of the section.

\subsection{Embedding and Bidirectional LSTM}

Given a $n$-word sentence $c=\{w_1^c,w_2^c,\cdots,w_{\tau+1}^c\\,\cdots,w_{\tau+m}^c,\cdots,w_{n-1}^c,w_n^c\}$ containing a corresponding $m$-word aspect starting from the $(\tau+1)$-th token, we embed each word token into a low-dimensional real-valued vector space~\cite{bengio2003neural} with embedding matrix $\mathbf{E}\in\mathbb{R}^{|V|\times d_e}$, where $|V|$ is the size of vocabulary and $d_e$ denotes the dimensionality of word embeddings. With the word embeddings of the sentence, a bidirectional LSTM is constructed to produce hidden state vectors $\mathbf{H}^c=\{\mathbf{h}_1^c,\mathbf{h}_2^c,\cdots,\mathbf{h}_{\tau+1}^{c},\cdots,\mathbf{h}_{\tau+m}^{c},\cdots,\mathbf{h}_{n-1}^c,\mathbf{h}_n^c\}$, where $\mathbf{h}_t^c\in\mathbb{R}^{2d_h}$ represents the hidden state vector at time step $t$ from the bidirectional LSTM, and $d_h$ is the dimensionality of a hidden state vector output by an unidirectional LSTM.

\subsection{Obtaining Aspect-oriented Features}

Different from general sentiment classification, aspect-based sentiment classification targets at judging sentiments from the view of aspects, and thus calls for an aspect-oriented feature extraction strategy. In this study, we obtain aspect-oriented features by applying multi-layer graph convolution over the syntactical dependency tree of a sentence, and imposing an aspect-specific masking layer on its top. 

\subsubsection{Graph Convolution over Dependency Trees}

Aiming to address the limitations of existing approaches (as discussed in previous sections), we leverage a graph convolutional network over dependency trees of sentences. Specifically, after the dependency tree\footnote{We use spaCy toolkit: \href{https://spacy.io/}{https://spacy.io/}.} of the given sentence is constructed, we first attain an adjacency matrix $\mathbf{A}\in\mathbb{R}^{n\times n}$ according to the words in the sentence. It is important to note that dependency trees are directed graphs. While GCNs generally do not consider directions, they could be adapted to the direction-aware scenario. Accordingly, we propose two variants of ASGCN, i.e. ASGCN-DG on dependency graphs which are un-directional, and ASGCN-DT concerning dependency trees which are directional. Practically, the only difference between ASGCN-DG and ASGCN-DT lies in their adjacency matrices: The adjacency matrix of ASGCN-DT is much more sparse than that of ASGCN-DG. Such setting is in accordance with the phenomenon that parents nodes are broadly influenced by their children nodes. Furthermore, following the idea of self-looping in~\citet{kipf2017semi}, each word is manually set adjacent to itself, i.e. the diagonal values of $\mathbf{A}$ are all ones.

The ASGCN variants are performed in a multi-layer fashion, on top of the bidirectional LSTM output in Section 3.1, i.e. $\mathbf{H}^{0} = \mathbf{H}^{c}$ to make nodes aware of context~\cite{zhang2018graph}. Then the representation of each node is updated with graph convolution operation with normalization factor~\cite{kipf2017semi} as below:
\begin{equation}
    \tilde{\mathbf{h}}_i^{l}=\sum_{j=1}^{n}\mathbf{A}_{ij}\mathbf{W}^l\mathbf{g}_j^{l-1}
\end{equation}
\begin{equation}
    \mathbf{h}_i^{l}={\rm ReLU}(\tilde{\mathbf{h}}_i^{l}/(d_i+1)+\mathbf{b}^l)
\end{equation}
where $\mathbf{g}_j^{l-1}\in\mathbb{R}^{2d_h}$ is the $j$-th token's representation evolved from the preceding GCN layer while $\mathbf{h}_i^{l}\in\mathbb{R}^{2d_h}$ is the product of current GCN layer, and $d_i=\sum_{j=1}^{n}\mathbf{A}_{ij}$ is degree of the $i$-th token in the tree. The weights $\mathbf{W}^{l}$ and bias $\mathbf{b}^l$ are trainable parameters.

It is worth noting that we do not have $\mathbf{h}_i^{l}$ immediately fed into successive GCN layer, but conduct a position-aware transformation in the first place:
\begin{equation}
    \mathbf{g}_i^{l}=\mathcal{F}(\mathbf{h}_i^{l})
\end{equation}
where $\mathcal{F}(\cdot)$ is a function assigning position weights, widely adopted by previous works~\cite{li2018transformation,tang2016aspect,chen2017recurrent}, for augmenting the importance of context words close to the aspect. By doing so we aim at reducing the noise and bias that may have naturally arisen from the dependency parsing process. Specifically, the function $\mathcal{F}(\cdot)$ is:
\begin{equation}
q_i=\begin{cases}
1-\frac{\tau+1-i}{n}& 1 \leq i< \tau+1 \\
0& \tau+1 \leq i \leq \tau+m \\
1-\frac{i-\tau-m}{n}& \tau+m < i \leq n
\end{cases}
\end{equation}
\begin{equation}
    \mathcal{F}(\mathbf{h}_i^{l})=q_i \mathbf{h}_i^{l}
\end{equation}
where $q_i\in\mathbb{R}$ is the position weight to $i$-th token. The final outcome of the $L$-layer GCN is $\mathbf{H}^{L}=\{\mathbf{h}_1^{L},\mathbf{h}_2^{L},\cdots,\mathbf{h}_{\tau+1}^{L},\cdots,\mathbf{h}_{\tau+m}^{L},\cdots,\mathbf{h}_{n-1}^{L},\mathbf{h}_{n}^{L}\}, \\ \mathbf{h}_{t}^{L}\in\mathbb{R}^{2d_h}$.

\subsubsection{Aspect-specific Masking}

In this layer, we mask out hidden state vectors of non-aspect words and keep the aspect word states unchanged:
\begin{equation}
    \mathbf{h}_t^{L} = \mathbf{0} \quad 1\leq t<\tau+1, \tau+m<t\leq n
\end{equation}

The outputs of this zero-masking layer are the aspect-oriented features $\mathbf{H}_{\rm mask}^{L}=\{\mathbf{0},\cdots,\mathbf{h}_{\tau+1}^{L},\\ \cdots,\mathbf{h}_{\tau+m}^{L},\cdots,\mathbf{0}\}$. Through graph convolution, these features $\mathbf{H}_{\rm mask}^{L}$ have perceived contexts around the aspect in such a way that considers both syntactical dependencies and long-range multi-word relations. 

\subsection{Aspect-aware Attention}

Based on the aspect-oriented features, a refined representation of the hidden state vectors $\mathbf{H}^{c}$ is produced via a novel retrieval-based attention mechanism. The idea is to retrieve significant features that are semantically relevant to the aspect words from the hidden state vectors, and accordingly set a retrieval-based attention weight for each context word. In our implementation, the attention weights are computed as below:
\begin{equation}
    \label{eq8}
    \beta_t=\sum_{i=1}^{n}\mathbf{h}_t^{c\top}\mathbf{h}_i^L=\sum_{i=\tau+1}^{\tau+m}\mathbf{h}_t^{c\top}\mathbf{h}_i^L
\end{equation}
\begin{equation}
    \alpha_t=\frac{{\rm exp}(\beta_t)}{\sum_{i=1}^{n}{\rm exp}(\beta_i)}
\end{equation}
Here, the dot product is used to measure the semantic relatedness between aspect component words and words in the sentence so that aspect-specific masking, i.e. zero masking, could take effect as shown in Equation~\ref{eq8}. The final representation for prediction is therefore formulated as:
\begin{equation}
    \mathbf{r}=\sum_{t=1}^{n}\alpha_t\mathbf{h}_t^c
\end{equation}

\subsection{Sentiment Classification}

Having obtained the representation $\mathbf{r}$, it is then fed into a fully-connected layer, followed by a softmax normalization layer to yield a probability distribution $\mathbf{p}\in\mathbb{R}^{d_p}$ over polarity decision space:
\begin{equation}
    \mathbf{p}={\rm softmax}(\mathbf{W}_p \mathbf{r}+\mathbf{b}_p)
\end{equation}
where $d_p$ is the same as the dimensionality of sentiment labels while $\mathbf{W}_p\in\mathbb{R}^{d_p\times2d_h}$ and $\mathbf{b}_p\in\mathbb{R}^{d_p}$ are the learned weight and bias, respectively.

\subsection{Training}
This model is trained by the standard gradient descent algorithm with the cross-entropy loss and $L_2$-regularization:
\begin{equation}
    {\rm Loss}=-\sum_{(c,\hat{p})\in C}\log \mathbf{p}_{\hat{p}} + \lambda \Vert \Theta \Vert_2
\end{equation}
where $C$ denotes the collection of data sets,  $\hat{p}$ is the label and $\mathbf{p}_{\hat{p}}$ means the $\hat{p}$-th element of $\mathbf{p}$, $\Theta$ represents all trainable parameters, and $\lambda$ is the coefficient of $L_2$-regularization. 


%% file: experiment.tex
\section{Experiments}

\subsection{Datasets and Experimental Settings}

Our experiments are conducted on five datasets: one (\textsc{Twitter}) is originally built by~\citet{dong2014adaptive} containing twitter posts, while the other four (\textsc{Lap14}, \textsc{Rest14}, \textsc{Rest15}, \textsc{Rest16}) are respectively from SemEval 2014 task 4~\cite{pontiki2014semeval}, SemEval 2015 task 12~\cite{pontiki2015semeval} and SemEval 2016 task 5~\cite{pontiki2016semeval}, consisting of data from two categories, i.e. laptop and restaurant. Following
previous work~\cite{tang2016aspect}, we remove samples with conflicting\footnote{An opinion target is associated with different sentiment polarities.} polarities or without explicit aspects in the sentences in \textsc{Rest15} and \textsc{Rest16}. The statistics of datasets are reported in Table~\ref{tab1}.

\begin{table}[]
\centering
\begin{tabular}{ccccc} 
\toprule
\multicolumn{2}{c}{Dataset} & \# Pos. & \# Neu. & \# Neg.  \\ 
\midrule
\multirow{2}{*}{\textsc{Twitter}}    & Train    & 1561                 & 3127                 & 1560                 \\ 
\cmidrule{2-5}
                            & Test     & 173                  & 346                  & 173                  \\
\midrule                            
\multirow{2}{*}{\textsc{Lap14}}     & Train    & 994                  & 464                  & 870                  \\ 
\cmidrule{2-5}
                            & Test     & 341                  & 169                  & 128                  \\ 
\midrule
\multirow{2}{*}{\textsc{Rest14}} & Train    & 2164                 & 637                  & 807                  \\ 
\cmidrule{2-5}
                            & Test     & 728                  & 196                  & 196                  \\ 
\midrule
\multirow{2}{*}{\textsc{Rest15}} & Train    & 912                 & 36                  & 256                  \\ 
\cmidrule{2-5}
                            & Test     & 326                  & 34                  & 182                  \\ 
\midrule
\multirow{2}{*}{\textsc{Rest16}} & Train    & 1240                 & 69                  & 439                  \\ 
\cmidrule{2-5}
                            & Test     & 469                  & 30                  & 117                  \\ 
\bottomrule
\end{tabular}
\caption{\label{tab1} Dataset statistics.}
\end{table}

For all our experiments, 300-dimensional pre-trained GloVe vectors~\cite{pennington2014glove} are used to initialize word embeddings. All model weights are initialized with uniform distribution. The dimensionality of hidden state vectors is set to 300. We use Adam as the optimizer with a learning rate of 0.001. The coefficient of $L_2$-regularization is 10\textsuperscript{−5} and batch size is 32. Moreover, the number of GCN layers is set to 2, which is the best-performing depth in pilot studies. 

The experimental results are obtained by averaging 3 runs with random initialization,  where Accuracy and Macro-Averaged F1 are adopted as the evaluation metrics. We also carry out paired t-test on both Accuracy and Macro-Averaged F1 to verify whether the improvements achieved by our models over the baselines are significant.

\subsection{Models for Comparison}

\begin{table*}[h]
\centering

\resizebox{\textwidth}{!}{
\begin{tabular}{lllllllllll}
\toprule
\multirow{2}{*}{Model}   & 
\multicolumn{2}{c}{\textsc{Twitter}} &
\multicolumn{2}{c}{\textsc{Lap14}} & 
\multicolumn{2}{c}{\textsc{Rest14}} &
\multicolumn{2}{c}{\textsc{Rest15}} &
\multicolumn{2}{c}{\textsc{Rest16}}  \\
\cmidrule{2-11}
                         & Acc. & F1              
                         & Acc. & F1               
                         & Acc. & F1 
                         & Acc. & F1 
                         & Acc. & F1 
                         \\
\midrule
SVM & 63.40\textsuperscript{$\sharp$} & 63.30\textsuperscript{$\sharp$} & 70.49\textsuperscript{$\natural$} & N/A & 80.16\textsuperscript{$\natural$} & N/A & N/A & N/A & N/A & N/A \\
LSTM & 69.56 & 67.70 & 69.28 & 63.09 & 78.13 & 67.47 &  77.37 & 55.17 & 86.80 & 63.88                 \\
MemNet & 71.48 & 69.90 & 70.64 & 65.17 & 79.61 & 69.64 & 77.31 & 58.28 & 85.44 & 65.99                 \\
AOA & 72.30 & 70.20 & 72.62 & 67.52 & 79.97 & 70.42 & 78.17 & 57.02 & 87.50 & 66.21                 \\
IAN & \textbf{72.50} & \textbf{70.81} & 72.05 & 67.38 & 79.26 & 70.09 & 78.54 & 52.65 & 84.74 & 55.21                 \\
TNet-LF & \textbf{72.98} & \textbf{71.43} & \textbf{74.61} & \textbf{70.14} & 80.42 & 71.03 & 78.47 & 59.47 & \textbf{89.07} & \textbf{70.43}                 \\
ASCNN & 71.05 & 69.45 & 72.62 & 66.72 & \textbf{81.73} & \textbf{73.10} & 78.47 & 58.90 & 87.39 & 64.56                 \\
\midrule
ASGCN-DT & 71.53 & 69.68 & 74.14\textsuperscript{$\dagger$} & 69.24\textsuperscript{$\dagger$} & \textbf{80.86}\textsuperscript{$\ddagger$} & \textbf{72.19}\textsuperscript{$\ddagger$} & \textbf{79.34}\textsuperscript{$\dagger \ddagger$} & \textbf{60.78}\textsuperscript{$\dagger \ddagger$} & 88.69\textsuperscript{$\dagger$} & 66.64\textsuperscript{$\dagger$}                 \\
ASGCN-DG & 72.15\textsuperscript{$\dagger$} & 70.40\textsuperscript{$\dagger$} & \textbf{75.55}\textsuperscript{$\dagger \ddagger$} & \textbf{71.05}\textsuperscript{$\dagger \ddagger$} & 80.77\textsuperscript{$\ddagger$} & 72.02\textsuperscript{$\ddagger$} & \textbf{79.89}\textsuperscript{$\dagger \ddagger$} & \textbf{61.89}\textsuperscript{$\dagger \ddagger$} & \textbf{88.99}\textsuperscript{$\dagger$} & \textbf{67.48}\textsuperscript{$\dagger$}                 \\
\bottomrule
\end{tabular}
}
\caption{\label{tab2} Model comparison results (\%). Average accuracy and macro-F1 score over 3 runs with random initialization. The best two results with each dataset are in bold. The results with $\natural$ are retrieved from the original papers and the results with $\sharp$ are retrieved from~\citet{dong2014adaptive}. The marker $\dagger$ refers $p<0.05$ by comparing with ASCNN in paired t-test and the marker $\ddagger$ refers $p<0.05$ by comparing with TNet-LF in paired t-test.}
\end{table*}

In order to comprehensively evaluate the two variants of our model, namely, ASGCN-DG and ASGCN-DT, we compare them with a range of baselines and state-of-the-art models, as listed below:
\begin{itemize}
    \setlength{\itemsep}{0pt}
    \setlength{\parsep}{0pt}
    \setlength{\parskip}{0pt}
    \item SVM~\cite{kiritchenko2014nrc} is the model which has won SemEval 2014 task 4 with conventional feature extraction methods.
    \item LSTM~\cite{tang2016effective} uses the last hidden state vector of LSTM to predict sentiment polarity.
    \item MemNet~\cite{tang2016aspect} considers contexts as external memories and benefits from a multi-hop architecture.
    \item AOA~\cite{huang2018aspect} borrows the idea of attention-over-attention from the field of machine translation.
    \item IAN~\cite{ma2017interactive} interactively models the relationships between aspects and their contexts.
    \item TNet-LF~\cite{li2018transformation} puts forward Context-Preserving Transformation (CPT) to preserve and strengthen the informative part of contexts.
\end{itemize}

In order to examine to what degrees GCN would outperform CNN, we also involve a model named ASCNN in the experiment, which replaces 2-layer GCN with 2-layer CNN in ASGCN\footnote{In order to ensure the length of input and output is consistent, kernel length is set to 3 and padding is 1.}.

\subsection{Results}

As is shown in Table~\ref{tab2},  ASGCN-DG consistently outperforms all compared models on \textsc{Lap14} and \textsc{Rest15} datasets, and achieves comparable results on \textsc{Twitter} and \textsc{Rest16} datasets compared with baseline TNet-LF and on \textsc{Rest14} compared with ASCNN. The results demonstrate the effectiveness of ASGCN-DG and the insufficiency of directly integrating syntax information into attention mechanism as in~\citet{he2018effective}. Meanwhile, ASGCN-DG performs better than ASGCN-DT by a large margin on \textsc{Twitter}, \textsc{Lap14}, \text{Rest15} and \textsc{Rest16} datasets. And ASGCN-DT's result is lower than TNet-LF's on \textsc{Lap14}. A possible reason is that the information from parents nodes is as important as that from children nodes, so treating dependency trees as directed graphs leads to information loss. Additionally, ASGCN-DG outperforms ASCNN on all datasets except \textsc{Rest14}, illustrating ASGCN is better at capturing long-range word dependencies, while to some extent ASCNN shows an impact brought by aspect-specific masking. We suspect \textsc{Rest14} dataset is not so sensitive to syntactic information. Moreover, the sentences from \textsc{Twitter} dataset are less grammatical, restricting the efficacy. We conjecture this is likely the reason why ASGCN-DG and ASGCN-DT get sub-optimal results on \textsc{Twitter} dataset.

\subsection{Ablation Study}

To further examine the level of benefit that each component of ASGCN brings to the performance, an ablation study is performed on ASGCN-DG. The results are shown in Table~\ref{tab3}. We also present the results of BiLSTM+Attn as a baseline, which uses two LSTMs for the aspect and the context respectively.

\begin{table*}[h]
\centering
\resizebox{\textwidth}{!}{
\begin{tabular}{lllllllllll}
\toprule
\multirow{2}{*}{Model}   & 
\multicolumn{2}{c}{\textsc{Twitter}} &
\multicolumn{2}{c}{\textsc{Lap14}} & 
\multicolumn{2}{c}{\textsc{Rest14}} &
\multicolumn{2}{c}{\textsc{Rest15}} &
\multicolumn{2}{c}{\textsc{Rest16}}  \\
\cmidrule{2-11}
                         & Acc. & F1              
                         & Acc. & F1               
                         & Acc. & F1 
                         & Acc. & F1 
                         & Acc. & F1 
                         \\
\midrule
BiLSTM+Attn & 71.24 & 69.55 & 72.83 & 67.82 & 79.85 & 70.03 & 78.97 & 58.18 & 87.28 & 68.18                    \\
ASGCN-DG & 72.15 & 70.40 & 75.55 & 71.05 & 80.77 & 72.02 & 79.89 & 61.89 & 88.99 & 67.48  \\ 
\midrule
ASGCN-DG w/o pos. & 72.69 & 70.59 & 73.93 & 69.63 & 81.22 & 72.94 & 79.58 & 61.55 & 88.04 & 66.63                 \\
ASGCN-DG w/o mask & 72.64 & 70.63 & 72.05 & 66.56 & 79.02 & 68.29 & 77.80 & 57.51 & 86.36 & 61.41                 \\
ASGCN-DG w/o GCN & 71.92 & 70.63 & 73.51 & 68.83 & 79.40 & 69.43 & 79.40 & 61.18 & 87.55 & 66.19                 \\
\bottomrule
\end{tabular}
}
\caption{\label{tab3} Ablation study results (\%). Accuracy and macro-F1 scores are the average value over 3 runs with random initialization.}
\end{table*}

First, removal of position weights (i.e. ASGCN-DG w/o pos.) leads to performance drops on \textsc{Lap14}, \textsc{Rest15} and \textsc{Rest16} datasets but performance boosts on \textsc{Twitter} and \textsc{Rest14} datasets. Recall the main results on \textsc{Rest14} dataset, we conclude that the integration of position weights is not helpful to reduce noise of user generated contents if syntax is not crucial for the data. Moreover, after we get rid of aspect-specific masking (i.e. ASGCN-DG w/o masking), the model could not keep as competitive as TNet-LF. This verifies the significance of aspect-specific masking.

Compared with ASGCN-DG, ASGCN-DG w/o GCN (i.e. preserving position weights and aspect-specific masking, but without using GCN layers) is much less powerful on all five datasets except F1 metric on \textsc{Twitter} dataset. However, ASGCN-DG w/o GCN is still slightly better than BiLSTM+Attn on all datasets except \textsc{Rest14} dataset, due to the strength of the aspect-specific masking mechanism. 

Thus it could be concluded that GCN contributes to ASGCN to a considerable extent since GCN captures syntatic word dependencies and long-range word relations at the same time. Nevertheless, the GCN does not work well as expected on the datasets not sensitive to syntax information, as we have seen in \textsc{Twitter} and \textsc{Rest14} datasets.  

\subsection{Case Study}

To better understand how ASGCN works, we present a case study with several testing examples.  Particularly, we visualize the attention scores offered by MemNet, IAN, ASCNN and ASGCN-DG in Table~\ref{tab4}, along with their predictions on these examples and the corresponding ground truth labels.

The first sample ``\textit{great food but the service was dreadful!}'' has two aspects within one sentence, which may hinder attention-based models from aligning the aspects with their relevant descriptive words precisely. The second sample sentence ``\textit{The staff should be a bit more friendly.}'' uses a subjunctive word ``\textit{should}'', bringing extra difficulty in detecting implicit semantics. The last example contains negation in the sentence, that can easily lead models to make wrong predictions.

MemNet fails in all three presented samples. While IAN is capable of differing modifiers for distinct aspects, it fails to infer sentiment polarities of sentences with special styles. Armed with position weights, ASCNN correctly predicts the label for the third sample as the phrase \textit{did not} is a significant signal for the aspect \textit{Windows 8}, but failed for the second one with a long-range word dependency. Our ASGCN-DG correctly handles all the three samples, implying that GCN effectively integrates syntactic dependency information into an enriched semantic representation. In particular, ASGCN-DG makes correct predictions on the second and the third sample, both having a seemingly biased focus. This shows ASGCN's capability of capturing long-range multi-word features.

\begin{table*}[h]
\centering
\begin{tabular}{cclcc}
\toprule
Model & Aspect & Attention visualization & Prediction & Label  \\
\midrule
\multirow{3}{*}[-0.5cm]{MemNet} & food & {\setlength{\fboxsep}{0pt}\colorbox{white!0}{\parbox{0.4\textwidth}{\colorbox{orange!51.521295042647175}{\strut great} \colorbox{orange!0.0}{\strut food} \colorbox{orange!51.6210970994153}{\strut but} \colorbox{orange!49.68677061797134}{\strut the} \colorbox{orange!50.7787277501318}{\strut service} \colorbox{orange!49.71496932488627}{\strut was} \colorbox{orange!100.0}{\strut dreadful} \colorbox{orange!49.803778118787065}{\strut !} 
}}} & negative\textsubscript{\xmark} & positive \\
\cmidrule{2-5}
& staff & {\setlength{\fboxsep}{0pt}\colorbox{white!0}{\parbox{0.4\textwidth}{\colorbox{orange!14.840717074960835}{\strut The} \colorbox{orange!0.0}{\strut staff} \colorbox{orange!43.20179219337782}{\strut should} \colorbox{orange!22.323401840987774}{\strut be} \colorbox{orange!18.617206434628113}{\strut a} \colorbox{orange!83.19593547232328}{\strut bit} \colorbox{orange!19.54690310752666}{\strut more} \colorbox{orange!100.0}{\strut friendly} \colorbox{orange!14.956261751285634}{\strut .} 
}}} & positive\textsubscript{\xmark} & negative \\
\cmidrule{2-5}
& Windows 8 & {\setlength{\fboxsep}{0pt}\colorbox{white!0}{\parbox{0.4\textwidth}{
\colorbox{orange!86.33507264142904}{\strut Did} \colorbox{orange!100.0}{\strut not} \colorbox{orange!92.13077664645047}{\strut enjoy} \colorbox{orange!28.167639762576673}{\strut the} \colorbox{orange!32.13807004013841}{\strut new} \colorbox{orange!0.0}{\strut Windows} \colorbox{orange!0.0}{\strut 8} \colorbox{orange!36.1855818278088}{\strut and} \colorbox{orange!60.88522180063477}{\strut touchscreen} \colorbox{orange!31.184652830797923}{\strut functions} \colorbox{orange!27.91015366511346}{\strut .} 
}}} & positive\textsubscript{\xmark} & negative \\
\midrule
\multirow{3}{*}[-0.5cm]{IAN} & food & {\setlength{\fboxsep}{0pt}\colorbox{white!0}{\parbox{0.4\textwidth}{\colorbox{orange!100.0}{\strut great} \colorbox{orange!99.90304110222749}{\strut food} \colorbox{orange!0.03803960492745688}{\strut but} \colorbox{orange!0.0007848750156963303}{\strut the} \colorbox{orange!0.0007211458477560407}{\strut service} \colorbox{orange!0.0}{\strut was} \colorbox{orange!0.0}{\strut dreadful} \colorbox{orange!10.256696392633929}{\strut !} 
}}} & positive\textsubscript{\cmark} & positive \\
\cmidrule{2-5}
& staff & {\setlength{\fboxsep}{0pt}\colorbox{white!0}{\parbox{0.4\textwidth}{\colorbox{orange!0.8404287095101841}{\strut The} \colorbox{orange!95.74960172498223}{\strut staff} \colorbox{orange!10.270666540389547}{\strut should} \colorbox{orange!0.1198812233732651}{\strut be} \colorbox{orange!0.0}{\strut a} \colorbox{orange!44.049399097459876}{\strut bit} \colorbox{orange!0.2906162251266379}{\strut more} \colorbox{orange!99.97401441331249}{\strut friendly} \colorbox{orange!100.0}{\strut .} 
}}} & positive\textsubscript{\xmark} & negative \\
\cmidrule{2-5}
& Windows 8 & {\setlength{\fboxsep}{0pt}\colorbox{white!0}{\parbox{0.4\textwidth}{
\colorbox{orange!2.6054923364442386}{\strut Did} \colorbox{orange!31.51981330749139}{\strut not} \colorbox{orange!0.09749540120300389}{\strut enjoy} \colorbox{orange!0.0}{\strut the} \colorbox{orange!0.011104271268752412}{\strut new} \colorbox{orange!9.222089748450307}{\strut Windows} \colorbox{orange!10.617286892244737}{\strut 8} \colorbox{orange!6.5807410213496595}{\strut and} \colorbox{orange!71.21417992820277}{\strut touchscreen} \colorbox{orange!94.6329925241964}{\strut functions} \colorbox{orange!100.0}{\strut .} 
}}} & neutral\textsubscript{\xmark} & negative \\
\midrule
\multirow{3}{*}[-0.5cm]{ASCNN} & food & {\setlength{\fboxsep}{0pt}\colorbox{white!0}{\parbox{0.4\textwidth}{
\colorbox{orange!100.0}{\strut great} \colorbox{orange!2.0486278067196375}{\strut food} \colorbox{orange!2.860578618878191}{\strut but} \colorbox{orange!0.0}{\strut the} \colorbox{orange!6.47838348183198}{\strut service} \colorbox{orange!2.0312273459677286}{\strut was} \colorbox{orange!7.427179255690477}{\strut dreadful} \colorbox{orange!89.81128041510085}{\strut !} 
}}} & positive\textsubscript{\cmark} & positive \\
\cmidrule{2-5}
& staff & {\setlength{\fboxsep}{0pt}\colorbox{white!0}{\parbox{0.4\textwidth}{
\colorbox{orange!0.26257849369440556}{\strut The} \colorbox{orange!0.4744346421778328}{\strut staff} \colorbox{orange!85.89984895966502}{\strut should} \colorbox{orange!100.0}{\strut be} \colorbox{orange!2.229805608491982}{\strut a} \colorbox{orange!51.41961417711018}{\strut bit} \colorbox{orange!15.789449991587478}{\strut more} \colorbox{orange!1.6008074150120348}{\strut friendly} \colorbox{orange!0.0}{\strut .} 
}}} & neutral\textsubscript{\xmark} & negative \\
\cmidrule{2-5}
& Windows 8 & {\setlength{\fboxsep}{0pt}\colorbox{white!0}{\parbox{0.4\textwidth}{
\colorbox{orange!0.015894787102755662}{\strut Did} \colorbox{orange!7.202180602614458}{\strut not} \colorbox{orange!0.010125839539040463}{\strut enjoy} \colorbox{orange!0}{\strut the} \colorbox{orange!0}{\strut new} \colorbox{orange!0.0005530223304343871}{\strut Windows} \colorbox{orange!0.0010989017160139192}{\strut 8} \colorbox{orange!0.0022465836650327957}{\strut and} \colorbox{orange!100.0}{\strut touchscreen} \colorbox{orange!0.9415278715874227}{\strut functions} \colorbox{orange!0.0}{\strut .}  
}}} & negative\textsubscript{\cmark} & negative \\
\midrule
\multirow{3}{*}[-0.5cm]{ASGCN-DG} & food & {\setlength{\fboxsep}{0pt}\colorbox{white!0}{\parbox{0.4\textwidth}{\colorbox{orange!100.0}{\strut great} \colorbox{orange!1.5357627869333699}{\strut food} \colorbox{orange!1.3871524043043946}{\strut but} \colorbox{orange!0.0}{\strut the} \colorbox{orange!0.018210808880066445}{\strut service} \colorbox{orange!0.3217879086517814}{\strut was} \colorbox{orange!46.61256684512336}{\strut dreadful} \colorbox{orange!6.3607324662109415}{\strut !} 
}}} & positive\textsubscript{\cmark} & positive \\
\cmidrule{2-5}
& staff & {\setlength{\fboxsep}{0pt}\colorbox{white!0}{\parbox{0.4\textwidth}{\colorbox{orange!18.891617978991626}{\strut The} \colorbox{orange!49.597193323443314}{\strut staff} \colorbox{orange!34.758832107631754}{\strut should} \colorbox{orange!9.676968222892565}{\strut be} \colorbox{orange!5.661785056448778}{\strut a} \colorbox{orange!100.0}{\strut bit} \colorbox{orange!9.768899448667575}{\strut more} \colorbox{orange!6.208355575599382}{\strut friendly} \colorbox{orange!0.0}{\strut .} 
}}} & negative\textsubscript{\cmark} & negative \\
\cmidrule{2-5}
& Windows 8 & {\setlength{\fboxsep}{0pt}\colorbox{white!0}{\parbox{0.4\textwidth}{
\colorbox{orange!0.3106986344594186}{\strut Did} \colorbox{orange!100.0}{\strut not} \colorbox{orange!0.5782619044935986}{\strut enjoy} \colorbox{orange!0.00015731715193376022}{\strut the} \colorbox{orange!0.0}{\strut new} \colorbox{orange!0.0016153174702502246}{\strut Windows} \colorbox{orange!0.0006360796391717312}{\strut 8} \colorbox{orange!0.04781964256067773}{\strut and} \colorbox{orange!58.08482958326475}{\strut touchscreen} \colorbox{orange!0.5718443483601369}{\strut functions} \colorbox{orange!0}{\strut .} 
}}} & negative\textsubscript{\cmark} & negative \\
\bottomrule
\end{tabular}
\caption{\label{tab4} Case study. Visualization of attention scores from MemNet, IAN, ASCNN and ASGCN-DG on testing examples, along with their predictions and correspondingly, golden labels. The marker\ \cmark\  indicates correct prediction while the marker\ \xmark\ indicates incorrect prediction.}
\end{table*}

%% file: discussion.tex
\section{Discussion}

\subsection{Investigation on the Impact of GCN Layers}

\begin{figure}[h]
    \centering
    \includegraphics[scale=0.5]{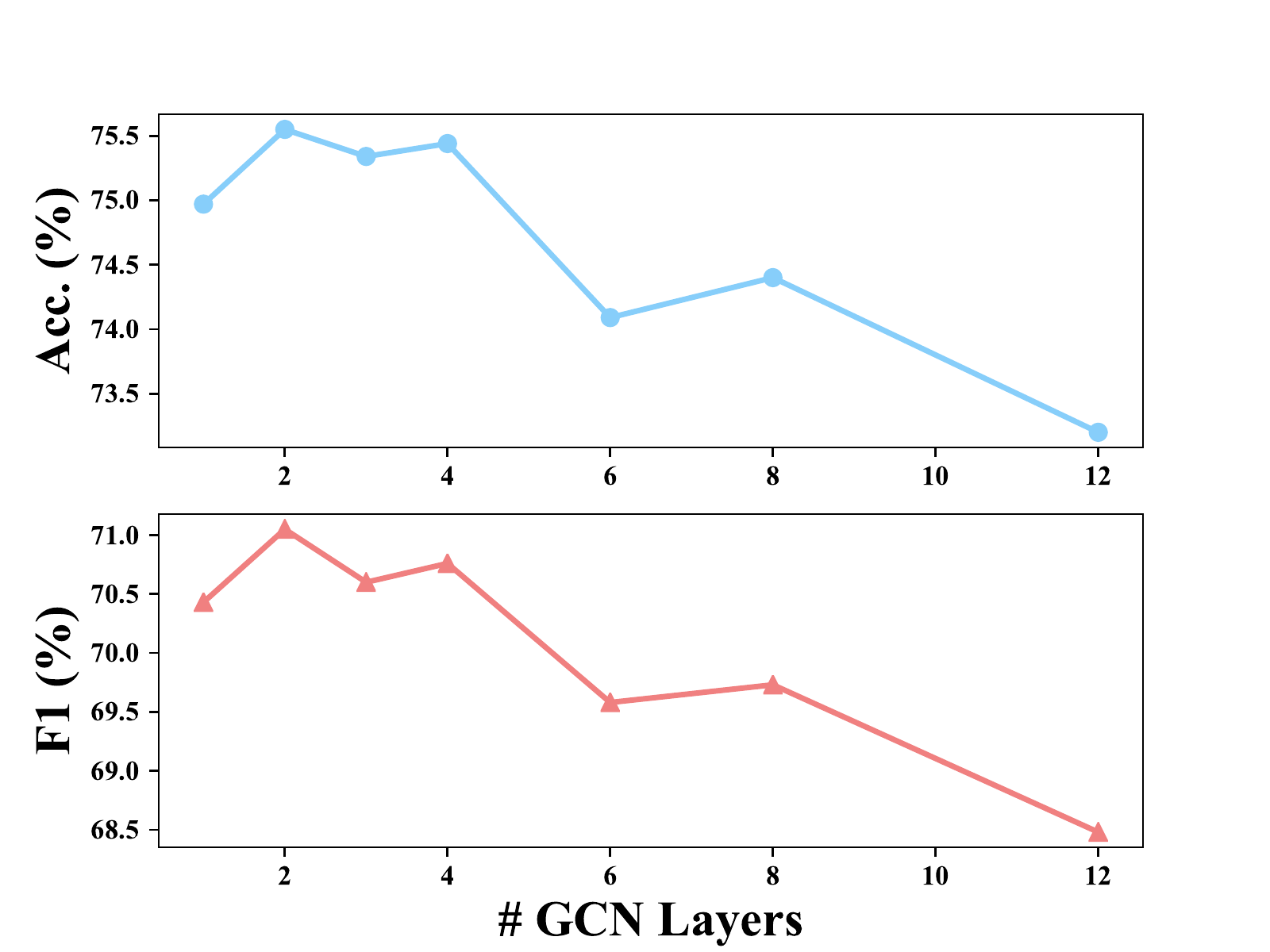}
    \caption{\label{fig3} Effect of the number of GCN layers. Accuracy and macro-F1 scores are the average value over 3 runs with random initialization.}
\end{figure}

As ASGCN involves an $L$-layer GCN, we investigate the effect of the layer number $L$ on the final performance of ASGCN-DG.  Basically, we vary the value of $L$ in the set \{1,2,3,4,6,8,12\} and check the corresponding Accuracy and Macro-Averaged F1 of ASGCN-DG on the \textsc{Lap14} dataset. The results are illustrated in Figure~\ref{fig3}.

On both metrics, ASGCN-DG achieves the best performance when $L$ is 2, which justifies the selection on the number of layers in the experiment section. Moreover, a dropping trend on both metrics is present as $L$ increases. For large $L$, especially when $L$ equals to 12, ASGCN-DG basically becomes more difficult to train due to large amount of parameters.

\subsection{Investigation on the Effect of Multiple Aspects}

In the datasets, there might exist multiple aspect terms in one sentence. Thus, we intend to measure whether such phenomena would affect the effectiveness of ASGCN. We divide the training samples in \textsc{Lap14} and \textsc{Rest14} datasets into different groups based on the number of aspect terms in the sentences and compute the training accuracy differences between these groups. It is worth noting that the samples with more than 7 aspect terms are removed as outliers because the sizes of these samples are too small for any meaningful comparison.

\begin{figure}[h]
    \centering
    \includegraphics[scale=0.5]{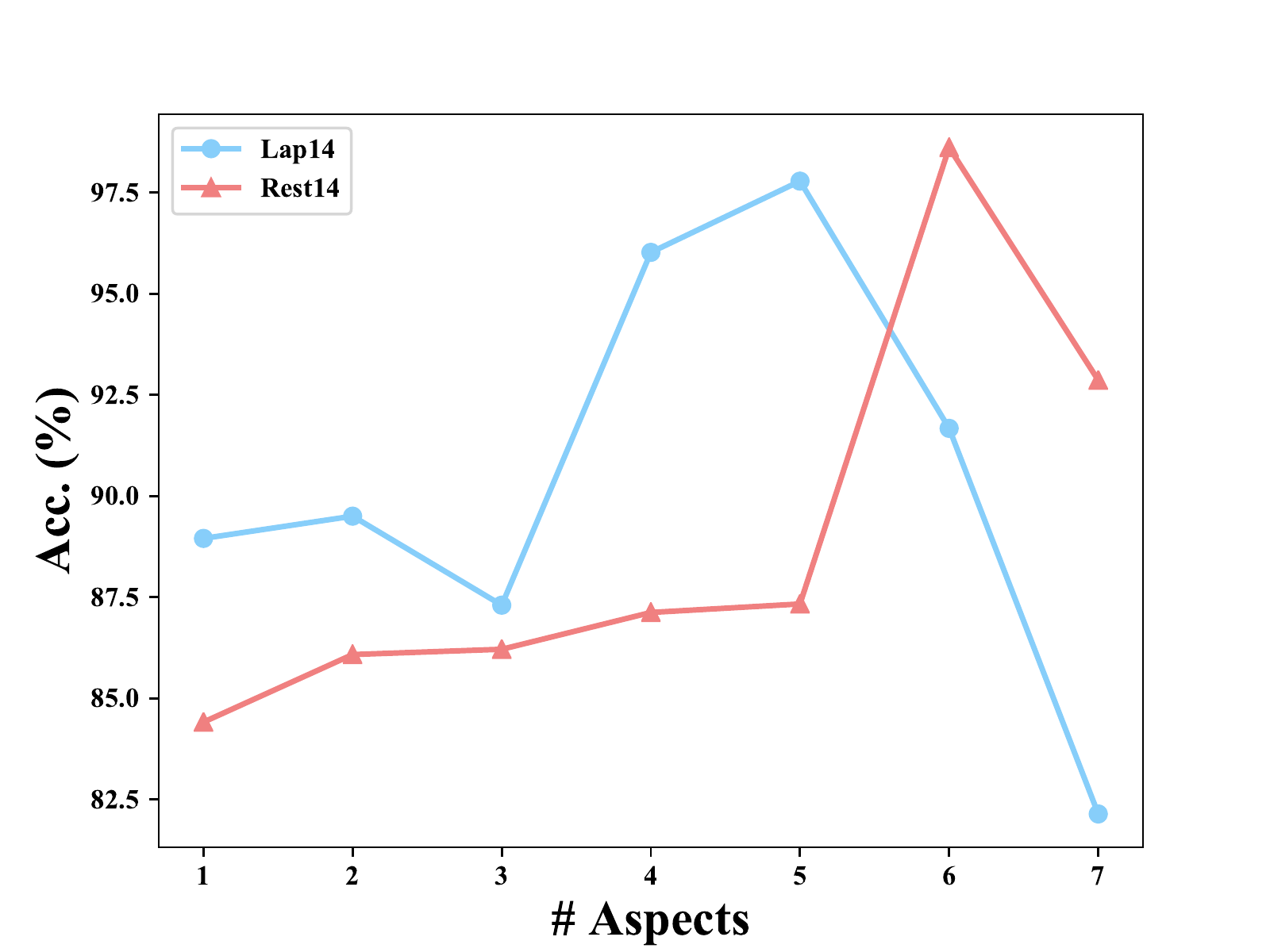}
    \caption{\label{fig4} Accuracy versus the number of aspects (\# Aspects) in the sentences.}
\end{figure}

It can be seen in Figure~\ref{fig4} that when the number of aspects in the sentences is more than 3, the accuracy becomes fluctuated, indicating a low robustness in capturing multiple-aspect correlations and suggesting the need of modelling multi-aspect dependencies in future work.

%% file: related.tex
\section{Related Work}

Constructing neural network models over word sequences, such as CNNs~\cite{kim2014convolutional,johnson2015semi}, RNNs~\cite{tang2016effective} and Recurrent Convolutional Neural Networks (RCNNs)~\cite{lai2015recurrent}, has achieved promising performances in sentiment analysis. However, the importance but lack of an effective mechanism of leveraging dependency trees for capturing distant relations of words has also been recognized. ~\citet{tai2015improved} showed that LSTM with dependency trees or constituency trees outperformed CNNs. ~\citet{dong2014adaptive} presented an adaptive recursive neural network using dependency trees, which achieved competitive results compared with strong baselines. More recent research showed that general dependency-based models are difficult to achieve comparable results to the attention-based models, as dependency trees are not capable of catching long-term contextualized semantic information properly. Our work overcomes this limitation by adopting Graph convolutional networks (GCNs) ~\cite{kipf2017semi} .

GCN has recently attracted a growing attention in the area of artificial intelligence and has been applied to Natural Language Processing (NLP). ~\citet{marcheggiani2017encoding} claimed that GCN could be considered as a complement to LSTM, and proposed a GCN-based model for semantic role labeling. ~\citet{vashishth2018dating} and ~\citet{zhang2018graph} used graph convolution over dependency trees in document dating and relation classification, respectively. ~\citet{yao2018graph} introduced GCN to text classification utilizing document-word and word-word relations, and gained improvements over various state-of-the-art methods. Our work investigates the effect of dependency trees in depth via graph convolution, and develops aspect-specific GCN model that integrates with the LSTM architecture and attention mechanism for more effective aspect-based sentiment classification.

%% file: conclusion.tex
\section{Conclusions and Future Work}

We have re-examined the challenges encountering existing models for aspect-specific sentiment classification, and pointed out the suitability of graph convolutional network (GCN) for tackling these challenges. Accordingly, we have proposed a novel network to adopt GCN for aspect-based sentiment classification. Experimental results have indicated that GCN brings benefit to the overall performance by leveraging both syntactical information and long-range word dependencies.

This study may be further improved in the following aspects. First, the edge information of the syntactical dependency trees, i.e. the label of each edge, is not exploited in this work. We plan to design a specific graph neural network that takes into consideration the edge labels. Second, domain knowledge can be incorporated. Last but not least, the ASGCN model may be extended to simultaneously judge sentiments of multiple aspects by capturing dependencies between the aspects words.